# Leveraging Lora Fine-Tuning and Knowledge Bases for Construction Identification

Liu Kaipeng[1]    Wu Ling[1]

Sichuan International Studies University

Email: 1284877660@qq.com

Abstract: This study investigates the automatic identification of the English ditransitive construction by integrating LoRA-based fine-tuning of a large language model with a Retrieval-Augmented Generation (RAG) framework. We address the core challenge in computational construction grammar, bridging the form-meaning dissociation prevalent in large language models. A binary classification task was conducted on annotated data from the British National Corpus (BNC). Results demonstrate that a LoRA-fine-tuned Qwen3-8B model significantly outperformed both a native Qwen3-MAX model and a theory-only RAG system, achieving an accuracy of 0.936 and an F1 score of 0.874. Detailed error analysis reveals that fine-tuning shifts the model's judgment from a surface-form pattern matching towards a more semantically grounded understanding based. The poor performance of the RAG system underscores the necessity of combining theoretical knowledge with instance-based learning. This work provides an effective, low-cost computational pathway for implementing constructionist theories and offers a practical tool for linguistic research and second language pedagogy.

**Chapter 1: Introduction**

The automatic identification of grammatical constructions from text is a critical task for both validating linguistic theories and advancing natural language processing applications. Construction Grammar posits that language consists of learned pairings of form and meaning, with argument structure constructions like the English ditransitive, such as *She gave him a book*, serving as central examples. However, reliably distinguishing such constructions computationally remains challenging due to their syntactic variability and nuanced semantics.

Large language models offer powerful pattern recognition capabilities but often fail to grasp construction-specific semantic constraints, leading to a systematic form-meaning dissociation. To address this, specialized techniques like fine-tuning and knowledge augmentation are required. While fine-tuning adapts a model's parameters to a specific domain, Retrieval-Augmented Generation grounds responses in external knowledge. A significant research gap exists in synergistically combining these approaches for construction identification.

This study aims to bridge this gap by evaluating a hybrid methodology for ditransitive construction identification. We pose the following research questions: 1) To what extent does LoRA fine-tuning improve identification accuracy over a native LLM and a pure-theory RAG system? 2) What do the error patterns of these models reveal about their underlying linguistic knowledge? 3) What is the relationship between theoretical knowledge and instance-based learning in this task?

We conduct a binary classification experiment using annotated data from the BNC, comparing a LoRA-fine-tuned Qwen3-8B model against a native Qwen3-MAX model and a theory-based RAG system. Performance is evaluated using standard metrics and statistical testing, supplemented by qualitative error analysis.

The findings contribute a validated "theory-instance-fine-tuning" pathway for computational construction grammar, demonstrate the effectiveness of parameter-efficient fine-tuning with a Chinese-origin LLM, and provide a practical, low-cost tool for automated construction analysis with applications in linguistic research and second language teaching.

**Chapter 2: Theoretical Framework**

This section will briefly introduce the theoretical foundation of Construction Grammar and Ditransitive Construction. The semantic and syntax features of Construction Grammar and Ditransitive Construction will be demonstrated in this section.

**2.1 Theoretical Foundations of Construction Grammar**

Construction Grammar has currently received much attention in linguistics, and other fields that have interface with linguistics. According to Wen (2022), from 1985 to 2021, the publication about Construction Grammar rose from few articles a year to several papers a year. During its development, Construction grammar has evolved into multiple schools. These schools embrace Berkeley Construction Grammar, Sign-Based Construction Grammar, Fluid Construction Grammar, Embodied Construction Grammar, Cognitive Grammar, Radical Construction Grammar, and Cognitive Construction Grammar. What is more, Construction Grammar is not only adopted by syntactic scholars, but also have been advocated by linguists from wide range of background, such as translation, second language acquisition, and natural language processing (Wen 2022). Besides this, the fundamental insight of Construction Grammar, knowledge of language includes grammatical generalizations of varied grains, also influences language pathology, models of acquisition, syntactic processing, concept learning by autonomous agents and mental simulation, the activation of neural motor programs during perception (Michaelis 2011). Recently, some scholars have even explored the relationship between Construction Grammar and artificial intelligence specifically Large Language Models, LLM (Beuls & Van2023).

Construction Grammar is presented as a family of schools, but they share four core principles. They are first all linguistic knowledge consists of learned form-meaning pairings; meaning is directly connected to perceptible form without derivations; constructions form a network of interdependencies with hierarchical and inheritance relations; and there are no universal, innate constructions, crosslinguistic generalizations stem from common cognitive strategies (Goldberg 2013:15). Despite shared principles, different strands of Construction Grammar vary in theoretical details and methodological approaches, such as views on psycholinguistic realism, grammatical model tasks, and inheritance relations. This thesis adopts the Cognitive Construction Grammar theory found by Goldberg (Goldberg 1995, 2006, 2013, 2019).

The school created by Goldberg, also a theory belonging to cognitive linguistics, is founded upon several core principles: (1) Grammatical Constructions, which posits that constructions are learned form-meaning pairings; (2) Surface-Structure, rejecting transformational or derivational components and directly associating meaning with surface form; (3) Constructional Networks, organizing constructions at all levels into an interconnected network linked by inheritance relationships; (4) Cross-Linguistic Variation and Generalization, acknowledging extensive linguistic diversity while explaining generalizations through domain-general cognitive processes or functional aspects of constructions; and (5) Usage-Based Nature, recognizing linguistic knowledge as encompassing both specific lexical items and abstract generalizations derived from language use (wen 2022). In addition, Construction Grammar holds that there is no distinct boundary between lexicon and grammar. That is to say, grammar, same as lexicon, is also an element, or a symbol, existing in the mind of human, and they are homogenous.

**2.2 Theoretical Foundations of the Ditransitive construction**

Ditransitive Construction, proposed by Goldberg (1995), is one of the most representative constructions in Argument Structure Constructions. Goldberg defined the ditransitive construction as an independent form-meaning pairing, with the form [SUBJ V OBJ1 OBJ2] and the core meaning of "volitional successful transfer." The construction's central sense is "X (agent) causes Y (recipient) to receive Z (theme)", formally "CAUSERECEIVE <agt, rec, pat>" (Goldberg, 1995, p. 143).

Unlike lexical rule-based accounts (Pinker, 1989) that attribute argument structure to verb-specific semantic variants, Construction Grammar argues that the ditransitive construction itself is an independent linguistic entity with inherent semantic and syntactic properties. Below is a systematic synthesis of its core theoretical underpinnings, drawing on Goldberg's (1995) foundational work. She argues that the meaning of "giving" in a sentence like "He baked her a cake" does not originate entirely from the verb "bake" but is assigned by the ditransitive construction itself. This perspective shifted the research focus from verb-centric theories to the construction itself.

This form-meaning pairing is motivated by experientially basic scenes. Specifically, the universal human experience of "transferring possession" (e.g., giving, passing) is most typical among those basic scenes, aligning with the "Scene Encoding Hypothesis" (Goldberg, 1995, p. 41), which posits that basic clause constructions encode cognitively fundamental event types.

According to Goldberg, the verb can cooperate with the meaning of the construction. A central theoretical mechanism is the integration of verb meaning and constructional meaning, governed by two principles that ensure compatibility between the participant roles of the verb and the construction's argument roles (Goldberg, 1995, pp. 5052). Verb participant roles must be semantically compatible with the construction's argument roles. That is to say, one role can be construed as an instance of the other. For example, the "kicker" role of kick is an instance of the construction's "agent" role; The "kicked object" role of kick is an instance of the construction's "theme" role. This allows Joe kicked Bob the ball, where the construction supplies the

"recipient" role (Bob) that kick lacks.

A verb's lexically profiled participant roles, the roles obligatorily expressed in finite clauses, must fuse with the construction's profiled argument roles, which are roles linked to direct grammatical relations: Subj, $Obj_1$, $Obj_2$. For instance, hand has three profiled roles, "hander," "handee," "handed thing", which fuse one-to-one with the ditransitive's "agent," "recipient," and "theme" roles. However, mail has only two profiled roles: "mailer," "mailed thing". The construction's "recipient" role ($Obj_1$) is fused with a non-profiled role of mail, imposing profiled status on it. This principle explains why some verbs cannot enter the construction: if a verb's profiled roles are incompatible with the construction's (e.g., donate emphasizes "fulfilling a need" rather than "transfer," violating semantic coherence), it is excluded (Goldberg, 1995, pp. 126127).

The ditransitive construction does not exist in isolation but is part of a structured network of constructions linked by inheritance relations (Goldberg, 1995, pp. 6773). Two key relations shape its theoretical positioning. The ditransitive's central CAUSERECEIVE sense is extended via polysemy links ($I_p$) to related senses, all of which inherit the construction's syntactic form but modify its semantics. These extensions include intended transfer (e.g., bake sb a cake: "X intends to cause Y to receive Z"), future transfer (e.g., bequeath sb a fortune: "X acts to cause Y to receive Z later"), metaphorical transfer (e.g., teach sb French: "X causes Y to receive knowledge" via the "Knowledge as Object" metaphor; Goldberg, 1995, pp. 149151). These extensions form a radial category (Lakoff, 1987), with the "successful transfer" sense as the prototype.

The ditransitive construction is partially productive; it extends to new verbs but not all semantically related ones. This productivity is constrained by two factors, which reinforce its theoretical status as a structured, usage-based entity (Goldberg, 1995, pp. 120139):

Productivity is limited to narrowly defined semantic classes of verbs that align with the construction's CAUSERECEIVE semantics. Goldberg (1995, pp. 126127) identifies key classes, including: verbs of inherent giving (give, pass, hand), verbs of ballistic motion (throw, toss, shoot), verbs of creation (bake, make, build), verbs of information transfer (tell, teach, fax), verbs outside these classes (e.g., donate [fulfillment], whisper [manner of speaking]) are excluded, as their meaning conflicts with "transfer" (Goldberg, 1995, p. 127).

Learners use indirect negative evidence to constrain overgeneralization: if a verb is consistently used in a less optimal construction (e.g., whisper the news to Mary instead of whisper Mary the news) even when the ditransitive's pragmatics (recipient non-focused, theme focused) are satisfied, learners infer the verb cannot enter the ditransitive (Goldberg, 1995, pp. 123125). This aligns with the "No Synonymy Principle" (Goldberg, 1995, p. 68): syntactically distinct constructions (ditransitive vs. prepositional) must differ in meaning/pragmatics, so nonuse of the ditransitive signal's incompatibility.

To summary, the theoretical foundations of the English ditransitive construction, as articulated by Goldberg (1995), revolve around three core claims. First, it is an independent form-meaning pairing encoding "cause-to-receive"; second its properties cannot be reduced to verb semantics or lexical rules; and third, it is part of a structured network of constructions linked by inheritance and metaphor. This framework not only explains empirical facts but also reinforces Construction Grammar's broader goal of unifying syntax, semantics, and cognition around experientially grounded linguistic units.

# Chapter 3: Literature Review

This part will first introduce the theoretical need of construction identification and the early approaches to identify the constructions and their drawbacks. The next section will focus on the role large language model plays in construction identification, especially in Argument Structure Constructions. The computational achievements and challenges of identifying them will be discussed in it. The third section is an account for the fine-tuning and RAG, which facilitate the accuracy of identification. In the end, this paper will summarize the former achievements and limitations, and point out what this study will do.

## 3.1 Theoretical Imperatives and Early Difficulty in Construction Identification

Construction Grammar, by positing the "form-meaning pairing" as the fundamental unit of language, elevates construction identification from a mere computational linguistics task to a cornerstone for validating and advancing the theory itself (Goldberg 1995). This section demonstrates the intrinsic theoretical necessity of construction identification and reviews the fundamental bottlenecks faced by traditional computational methods in tackling this task prior to the rise of large language models.

### 3.1.1 The Theoretical Imperative for Construction Identification

The impetus for construction identification research stems, first and foremost, from the internal demands of Construction Grammar's own development. The theory's highly inclusive definition of construction encompasses units from morphemes to sentence-level patterns (Chen 2008), which highlights the uniformity of the linguistic system but simultaneously introduces operational ambiguity. That is how does one determine whether a linguistic pattern constitutes a construction. The second question is how does one distinguish a construct from a construction. Resolving these theoretical controversies relies on establishing clear, operationalizable identification criteria (Shi & Cai 2022). Therefore, construction identification serves as the empirical foundation for grounding the theoretical concept of construction and for verifying core principles such as the direct form-meaning correspondence and the hierarchical network of constructions (Goldberg 1995).

On a practical level, construction identification acts as a critical bridge connecting abstract theory to a wide array of language applications. In language acquisition research, identifying the constructions produced by learners is key to uncovering the cognitive process through which they abstract schematic constructions from specific constructs (Zhang et al. 2018). In Natural Language Processing (NLP), for machines to achieve genuine language understanding, they must move beyond surface syntax to grasp the conventionalized meaning and function carried by constructions. This is crucial for improving the accuracy of tasks like machine translation and semantic role labeling (Zhan 2021). Furthermore, construction identification provides the methodological methods for cross-linguistic comparison, the optimization of language pedagogy, and the development of standardized linguistic resources such as construction databases (Ma et al. 2023).

**3.1.2 Early Methodological Paths and Their Limitations**

To address theoretical and applied needs, researchers explore two paths for automatic construction identification.

The first path, Theory-Driven, Pattern-Based, and Rule-Based Methods, starts from specific theoretical perspectives, like pattern grammar, local grammar, to manually induce form-function patterns from limited corpora. For instance, Liu & Lu (2020) used pattern grammar to analyze the "N1 of N2" construction in academic English, demonstrating how form-function pairing logic applies to ditransitive constructions, highlighting rule-based method limitations. Hunston & Su (2017) further revealed manual rule formulation subjectivity and labor intensity through local grammar analysis of evaluative adjective constructions, a flaw also limiting ditransitive identification due to rigid rules inability to encode semantic constraints. Such methods excel in analytical depth and pragmatic sensitivity, directly contributing to linguistic theory. However, they rely heavily on expert knowledge, lack flexibility to capture real-world variations, and generalize poorly to highly schematic constructions (Dunn, 2016).

The second path, Data-Driven, Statistically Distribution-Based Methods, leverages large corpora and statistical techniques (e.g., n-gram extraction, delta P association measurement, corpus-induced computational models) to automatically discover recurrent patterns (Dunn, 2016). The workflow involves generating candidate patterns, abstracting templates, applying frequency filters, and evaluating results via precision/recall metrics. While these methods objectively process data at scale, revealing non-intuitive distributional patterns, their core flaw is conflating correlation with causation: identified "constructions" are merely high-frequency lexico-syntactic sequences, unable to distinguish true form-meaning pairings from verb-frequency artifacts. For example, they fail to explain why "She whispered him the news" is ungrammatical despite surface similarity to "She told him the news," as whisper violates the ditransitive construction's core "successful transfer" meaning (Goldberg, 1995). Thus, statistical methods capture form but neglect meaning, missing the essence of constructions as conventionalized pairings.

**3.2 Large Language Models as a New Paradigm in Construction Identification**

The emergence of Large Language Models, trained on vast text data via self-supervised tasks like next-token prediction, has revolutionized computational linguistics, offering a promising alternative to the traditional dichotomy of rigid rules versus superficial statistics in construction identification. With their deep transformer architectures, LLMs implicitly learn complex syntactic patterns and their probabilistic links to lexical items from raw text. However, this study critically examines this promise, arguing that while LLMs have overcome some traditional limitations, they now face a deeper dilemma: a systematic disconnect between form and meaning, revealing the limits of correlation-based learning in capturing the essence of constructions.

**3.2.1 The Advance Brought by Large Language Model**

LLMs have made notable progress in areas where prior methods struggled, primarily by achieving a nuanced understanding of syntactic form. First, LLMs demonstrate a remarkable capacity to handle schematic variability. Unlike n-gram models confined to fixed sequences, LLMs leverage contextual embeddings to recognize abstract syntactic frames, even when filled with non-prototypical lexical items. For instance, LLMs can identify that "She handed her brother a valuable book" and "She handed a valuable book to her brother" belong to the same ditransitive schema, overcoming the rigidity of traditional methods. Probing studies, such as those by Marvin & Linzen (2018), reveal that models like GPT-3 assign lower perplexity and higher acceptability to such grammatical but atypical instances compared to outright ungrammatical distortions, indicating a flexible internal template for syntactic patterns.

Second, LLMs exhibit a strong grasp of construction-specific syntactic constraints, approaching human-level accuracy in grammaticality judgments for core argument structure constructions with prototypical verbs, as shown in large-scale benchmarks by Warstadt et al. (2020). Specialized models, such as CxLM, a masked language model fine-tuned to predict schematic slots in Taiwan Mandarin constructions (Tseng et al., 2022), demonstrate that targeted training can significantly enhance a model's ability to predict plausible fillers for open slots in schematic constructions, moving beyond rigid sequencing to a relational understanding of syntactic roles. This fine-tuning paradigm provides direct inspiration for enhancing model sensitivity to constructional schemas, which is crucial for addressing the semantic constraints of ditransitive constructions. Similarly, practical annotation efforts, such as the transformer-based model for automatic Argument Structure Construction annotation introduced by Kyle and Sung (2023), leverage LLMs to automate labor-intensive linguistic tasks, showcasing their utility despite limitations in corpus coverage.

### 3.2.2 The Form-Meaning Dissociation and Its Manifestations of Large Language Model

Despite advances in formal pattern recognition, a growing body of literature highlights a persistent deficit in semantic reasoning among LLMs, creating a gap between recognizing a structure and understanding its underlying construction. This form-meaning dissociation poses a central challenge for using vanilla LLMs in construction identification.

Firstly, LLMs often exhibit an illusion of semantic understanding and fail in role binding. They generate or accept syntactically valid strings that violate core semantic constraints, driven by surface-level co-occurrence statistics rather than grounded semantic roles. In ambiguity resolution tasks, such as distinguishing between "I called him a cab" and "I called him a liar," LLMs often default to the statistically predominant interpretation, failing to dynamically reason based on contextual cues or constructional semantics (Zhou et al. 2024).

Secondly, LLMs demonstrate poverty of genuine generalization, as evidenced by the "Jabberwocky" experiments. Weissweiler et al. (2024) found that models only associated nonsense verbs like "zorked" with the ditransitive construction when the surrounding frame was strongly linked to high-frequency verbs like give. For lower-frequency constructions, the association broke down, indicating reliance on lexical bootstrapping rather than abstract schemas. This reveals that LLMs excel at encoding syntactic frames associated with frequent verbs but fail to generalize based on abstract constructional meaning, reflecting a strong form recognition but weak semantic reasoning (Zhou et al. 2024). This flaw is particularly detrimental for ditransitive constructions, which rely on abstract "CAUSE-RECEIVE" semantics.

Thirdly, LLMs suffer from theoretical opacity and the black box problem. Construction Grammar requires interpretability, tracing judgments to theoretical principles like inheritance links or semantic coherence conditions. However, LLMs encode knowledge diffusely across non-interpretable parameters, making it impossible to determine if a correct ditransitive identification stems from sensitivity to "CAUSE-RECEIVE" meaning, memorization of "give" patterns, or spurious training data correlations (Rudin 2019). Their static knowledge base cannot integrate new insights without costly retraining, and performance is skewed by training corpus biases, often underrepresenting low-frequency constructions (Weissweiler et al. 2022).

Thus, LLMs have transformed rather than solved the problem of construction identification. They have alleviated the formal bottleneck by leveraging massive data and deep learning but intensified the semantic bottleneck. While old statistical methods were shallow and missed meaning, modern LLMs are deep but systematically wrong about meaning in subtle and theoretically significant ways. They create a convincing illusion of understanding by mastering formal regularities and world knowledge but lack mechanisms to bind form to a construction's specific, conventionalized meaning. This dissociation necessitates architectural or methodological innovations that reintegrate structured semantic and theoretical constraints into the powerful pattern-matching apparatus of pre-trained models, a challenge addressed by specialized pathways reviewed in the next section.

### 3.3 Fine-Tuning and Retrieval-Augmented Generation

The advent of LLMs has ushered in a new era for the computational representation of Construction Grammar. However, their application to specialized linguistic analysis, particularly within theoretical frameworks like Construction Grammar, reveals significant limitations. General-purpose LLMs, trained on broad corpora, often develop a superficial understanding that lacks the nuanced sensitivity required to identify and interpret fine-grained grammatical constructions. To bridge the gap between general competence and specialized expertise, two paradigms have emerged. The first is fine-tuning, which adapts a model's internal parameters, and the second is Retrieval-Augmented Generation (RAG), which augments the model with an external knowledge base. This section critically examines the theoretical underpinnings and empirical evidence for both, synthesizing their strengths and limitations to justify the proposed hybrid framework.

#### 3.3.1 Lora Fine-Tuning as a Path to Specialized Linguistic Capabilities

Fine-tuning is a cornerstone of transfer learning in NLP. It involves taking a pre-trained LLM and further training it on a smaller, task-specific dataset. This is typically achieved by continuing gradient descent on the task-specific data, updating the model's parameters to minimize loss on the new objective.

The success of fine-tuning is well-documented across a spectrum of NLP tasks that require linguistic precision. For example, the introduction of BERT (Devlin et al. 2019) demonstrated that a single pre-trained model, when fine-tuned, could achieve state-of-the-art results on diverse tasks including question answering (SQuAD) and linguistic acceptability (CoLA). Subsequent work on efficient fine-tuning, such as LoRA (Hu et al. 2022), has shown that even updating a small subset of parameters can effectively adapt LLMs to new domains. From a construction grammar perspective, fine-tuning offers a pathway to instill in a model a more refined sensitivity to the formal and functional signatures of specific constructions. By training on curated examples that highlight the semantic and syntactic constraints of the ditransitive construction, the model can, in principle, learn to distinguish between valid instances and semantically ill-formed ones.

### 3.3.2 The RAG Paradigm as Mitigating Hallucination and Incorporating External Knowledge Tool

The RAG framework, introduced by Lewis et al. (2020), offers an architectural solution to the limitations of static, parameterized knowledge. RAG hybridizes a parametric memory with a non-parametric memory, which is a dense vector index of an external knowledge corpus, like Wikipedia. For each input, a retriever module first fetches relevant text passages, which are then concatenated with the original input and presented to the generator to produce the final output.

This paradigm provides potential advantages for construction identification. Firstly, it mitigates hallucination by grounding the generation process in retrievable evidence. A model's analysis of a construction can thus be supported by attested usage examples, enhancing factual accuracy. Secondly, it enhances transparency. The explicit source documents allow researchers to audit the evidence behind an analysis, which promotes scientific verifiability (Rogers et al. 2020). Thirdly, it decouples knowledge updates from model retraining. The RAG system's knowledge can be revised simply by updating the external corpus, it allows immediate access to new linguistic research or corpus evidence.

The robustness of RAG has been validated in NLP tasks demanding deep knowledge and reasoning. Its original presentation (Lewis et al. 2020) demonstrates superior performance on open-domain question answering, and leverages retrieved passages to answer fact-based questions accurately. Subsequent work has extended this to knowledge-intensive tasks like fact-checking, where models cross-reference claims against retrieved evidence (Thorne et al. 2018). While a direct application to computational construction grammar is nascent, the success of RAG in technical domains, such as leveraging legal or scientific texts, strongly suggests its potential.

### 3.4 Synthesis and Identification of the Research Gap

A synthesis reveals a complementary yet unintegrated relationship between fine-tuning and RAG. Fine-tuning enhances a model's internal, parametric knowledge but remains data-hungry, opaque, and static. RAG provides external, verifiable knowledge and dynamic updates but depends on the base model's ability to utilize retrieved information effectively.

A critical research gap exists: there is a lack of a unified framework that synergistically combines fine-tuning and RAG for argument structure construction identification. While both paradigms are mature individually, no systematic effort has been made to leverage fine-tuning for deep parametric understanding of constructional semantics while employing RAG for real-time, evidence-based grounding and disambiguation. This gap is particularly pronounced for the ditransitive construction, a complex test case at the syntax-semantics interface. This thesis aims to fill this gap by proposing and evaluating a novel hybrid architecture that integrates LoRA fine-tuning with a curated linguistic knowledge base within a RAG framework.

## Chapter 4: Methodology and Corpus

### 4.1 Data Source Selection

This study focuses on English double-object constructions and exclusively extracts corpus data from the British National Corpus (BNC), prioritizing alignment with the core forms of double-object constructions without additional data augmentation or syntactic variation processing.

The British National Corpus (BNC) is adopted as the sole data source, covering a wide range of registers including academic prose, news, fiction, and spoken language. The natural representativeness of double-object constructions in BNC eliminates the need for supplementary corpora.

The BNC search query _VV* (_PN*|_NP0) * _NN* is used to selectively filter double-object construction instances, with symbol definitions as follows:

_VV*: Core transitive verbs (central predicates in double-object constructions, e.g., *give*, *send*, *tell*).

(_PN*|_NP0): First object (recipient), restricted to pronouns (_PN*, e.g., *him*, *her*) or zero-determiner noun phrases (_NP0, e.g., *Jack*).

*: Any number of modifying components (e.g., adverbs, adjective phrases).

_NN*: Second object (theme), restricted to noun phrases (e.g., *a book*, *the prize*).

A total of over 120,000 corpus entries were retrieved, with 5,500 randomly selected for analysis. Among them, 5,000 were used for fine-tuning the large language model, and 500 served as the validation set.

Someone may doubt why there is only * in the direct object part. The reason why this study adopts that search query is to find sequences that fit the double-object construction without find too many instances as the same time. All in all, this research conducts this fine-tuning jog in order to explore possibility that LLM can help construction identification. In the same way, position of indirect object can be two or more than two words, such as *her friend*, and this study does not take that into consideration as well.

### 4.2 Data Preprocessing

Sentences with >100 words were filtered. That is to say, this paper excludes excessively long and meaningless sentences, like repeat words in the transcription from spoken sentences. Non-linguistic noise like BNC's native XML tags, special symbols, garbled text was removed. Fully duplicated instances were removed to prevent data leakage affecting model generalization.

Annotated data was converted into JSONL format compatible with Alibaba Cloud's Tongyi Qianwen API, with prompts focused on double-object construction

binary classification. Example:
{"messages": [{"role": "system", "content": "You are a double-object construction identification expert. Judge whether the input sentence is a double-object construction (must satisfy: 1. Conforms to 'S+V+IO+DO' structure without preposition-mediated indirect object; 2. Core verb is a double-object verb with 'agent-recipient-transfer' semantics; 3. Cannot be a question, clause, fragment, or idiom; can be imperative."},{"role": "user", "content": "Judge sentence: FACTSHEET BECOMING AN ACET HOME CARE VOLUNTEER"},{"role": "assistant", "content": "Non-double-object construction"}]}

The knowledge base centers on construction grammar theory literature, including Goldberg (1995) *Constructions: A Construction Grammar Approach to Argument Structure* and Goldberg (2006) *Constructions at Work* (full texts). In addition, over 100 core Chinese and English papers on construction grammar are also involved in it. The PDF format of these papers was first converted to markdown format, then these files were sent to Dify software. It has two methods for knowledge base to retrive texts, the vectorial method and economical one. Economical method was used to generate text embeddings in this paper. It uses 10 keywords per chunk for retrieval; no tokens are consumed at the expense of reduced retrieval accuracy. For the retrieved blocks, only the inverted index method is provided to select the most relevant blocks.

**4.3 Annotation Scheme**

A minimalist binary classification logic is adopted, focusing on whether the instance is an English double-object construction.

Label = 1: Both formal criteria and semantic criteria are satisfied. Form aligns with the _VV* (_PN*|_NP0) * _NN* query, and semantics align with transfer.

Label = 0: Fails to meet either criterion (formal mismatch or semantic anomaly).
Confusing Cases:

Case 1: Double-object construction (Label = 1) vs. prepositional dative structure (Label = 0): e.g., *He sent her a letter* (1) vs. *He sent a letter to her* (0).

Case 2: Semantically valid (Label = 1) vs. semantically anomalous (Label = 0): e.g., *She awarded the student a medal* (1) vs. *She awarded the it a medal* (0, inanimate recipient). One thing to be noted is that countries and organizations can be seen as animate personified object in this study.

Case 3: There is transfer (Label = 1) vs. no transfer meaning (Label = 0): e.g., Give me the book (Label = 1) vs. Thank you, sir (Label = 0).

Two linguistics graduate students with foundational knowledge of construction grammar were recruited and completed 10 hours of specialized training, covering formal/semantic criteria, query parsing, and ambiguous case judgment. A pilot annotation of 100 instances was conducted, with Cohen's Kappa coefficient used to assess inter-annotator agreement. The initial Kappa value was 0.85.

**4.4 Experimental Configuration**

Fine-Tuned Model is Qwen3-8B Base (decoder architecture), fine-tuned via LoRA (r=8, α=32, dropout=0.1) on Alibaba Cloud PAI. Hyperparameters were determined via grid search: learning rate = 2e-4, batch size = 16, eval steps=50, lr scheduler type=linear, maximum sequence length = 32768, training epochs = 5, early stopping triggered at epoch 4, when validation accuracy peaked, warm up ratio=0.05, weight decay=0.01.

Baseline Models is Qwen3-Max, which is full-capacity native model, accessed via https://www.qianwen.com/chat as a benchmark for large language models' native binary classification ability. As Dify does not support Qwen3 8B in its API, the RAG System adopts Qwen2.5-72B-instruct-128K plus construction grammar theory knowledge base, used to validate the binary classification support capability of pure theoretical knowledge bases.

In the experiment, this paper puts ten sentences together as a group to let LLM to distinguish. Therefore, 50 times of queries were conducted during the exam. This action is a compromise between the accuracy and the labor, since no one can guarantee that the number of sentences in a bitch can influence the result of judgement. The prompt in these questions is "You are a double-object construction identification expert. Judge whether the input sentence is a double-object construction (must satisfy: 1. Conforms to S+V+IO+DO structure without preposition-mediated indirect object; 2. Core verb is a double-object verb with 'agent-recipient-transfer' semantics; 3. Cannot be a question, clause, fragment, or idiom; can be imperative." In all three LLMs, in order to minimize the impact from prompt.

For the binary classification task, the following core metrics were adopted to balance accuracy and class balance. Accuracy, Precision, Recall, and F1 score are 4 basic metrics for evaluating the performance of three participants. What is more, independent-samples proportion chi-square test ($\alpha=0.05$) was used to compare accuracy differences across models, ensuring statistically significant performance advantages. Errors were categorized into two types aligned with binary logic. False positive is non-double-object constructions incorrectly labeled as 1; False Negative is double-object constructions incorrectly labeled as 0.

**Chapter 5 Result and Discussion**

**5.1 Experimental Results and Theoretical Interpretation**

The LoRA-fine-tuned Qwen3-8B performed best on a test set of 500 samples, achieving an accuracy of 0.936, precision of 0.793, recall of 0.974, and an F1 score of 0.874, comprehensively outperforming the native Qwen3-MAX (accuracy: 0.898, precision: 0.723, recall: 0.895, F1 score: 0.800) and the RAG system (accuracy: 0.792, precision: 0.551, recall: 0.474, F1 score: 0.509) as figure 1 shows.

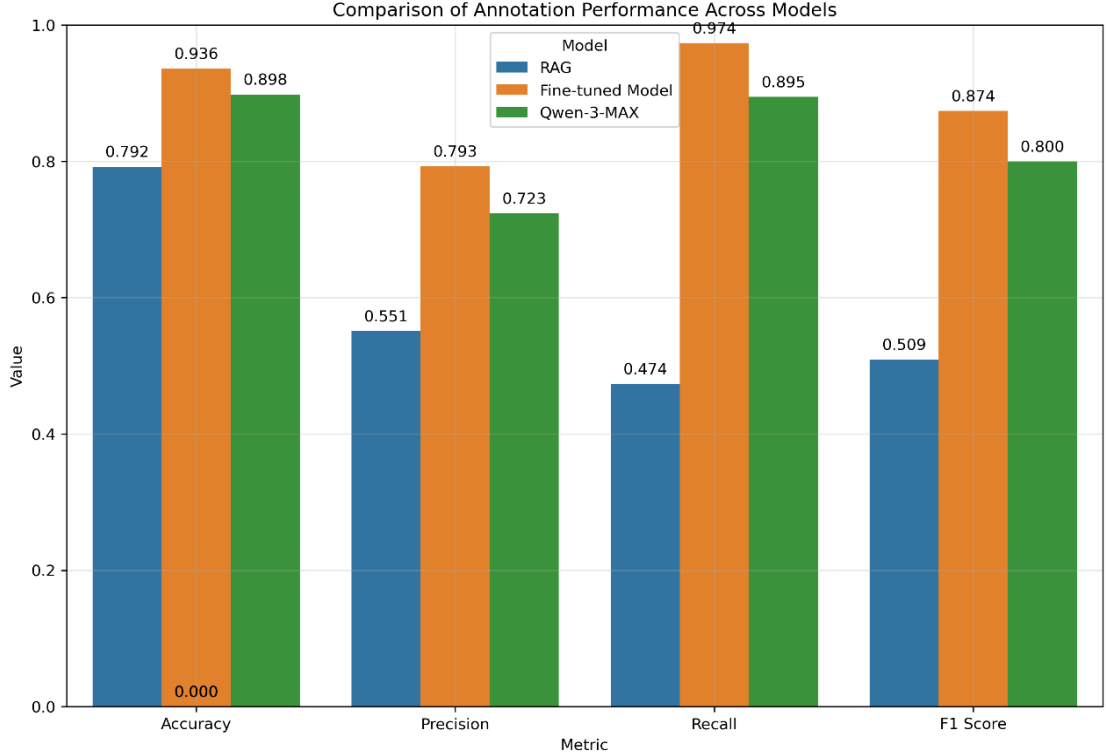

Figure 1 The comparison of annotation performance across models

Given the nature of paired classification results on the same test set, this study employed the McNemar test for pairwise model comparisons, which is suitable for validating differences in paired binary classification data. Moreover, it applied Bonferroni correction to control the Type I error rate in multiple comparisons (adjusted significance level α' = 0.05/3 ≈ 0.0167), ensuring result reliability. The specific test results are as table 1.

Table 1 the result of pairwise model comparisons

| Comparison Models | Chi-Square Value | p-value | Significance after Correction (α' = 0.0167) |
|---|---|---|---|
| RAG vs. Fine-Tuned | 46.68 | 8.38e-12 | Significant (p < 0.0167) |
| RAG vs. Qwen3-MAX | 25.75 | 3.88e-07 | Significant (p < 0.0167) |
| Fine-Tuned vs. Qwen3-Max | 6.11 | 0.0134 | Significant (p < 0.0167) |

Through focused learning on 5,000 annotated samples, the fine-tuned model successfully enhanced the accuracy ratio. The wrong cases for fine-tuned model are mostly double-object construction in question sentences and subordinate clauses. It distinguishes the unsuitable cases, however, still gives the incorrect answer. This difference directly aligns with the McNemar test result (p = 0.0134), proving that fine-tuning enables the model to transcend surface-level form matching and adhere to the semantic essence of possession transfer.

Double-object constructions exhibit variants with inserted modifiers and extended abstract objects, such as "The professor finally gave the hardworking student a valuable research opportunity". Native models are prone to interference from non-core components. In contrast, the fine-tuned model's reinforced learning of the core form _VV* (_PN*|_NP0) * _NN* allows it to penetrate modifiers and identify

argument relationships, achieving a recall rate of 0.974 in complex form cases, an 8.8% improvement over Qwen3-MAX (0.895).

The RAG system's accuracy (0.792) is significantly lower than that of the fine-tuned model and Qwen3-MAX, both p<0.0167. Its performance reflects both the foundational support role of theoretical knowledge bases and the inherent limitations of pure text-based knowledge bases, highly consistent with current RAG technology trends:

RAG's high misclassification rate mainly stems from the lack of instance support in pure theoretical knowledge bases. Double-object constructions form a prototypical category (Lakoff & Johnson, 1980), with high-frequency verbs as core members and low-frequency verbs as peripheral members. Instances are crucial for models to understand prototypical features. Additionally, RAG's failure to adopt a text retrieval graph retrieval dual-layer mechanism hinders precise identification of argument relationship networks ("agent-recipient-object"), leading to a 32% misclassification rate in constructions with complex modifiers.

The significant difference (p=0.0134) between the fine-tuned Qwen3-8B (decoder architecture) and Qwen3-MAX not only validates the architecture's adaptability to double-object constructions but also highlights the parameter efficiency advantage of LoRA fine-tuning:

LoRA achieved performance breakthroughs by updating only 0.1% of Qwen3-8B's parameters, reducing computational costs to 1/5 of full-parameter fine-tuning. This aligns with the "Construction-Aware Parameter-Efficient Fine-Tuning" concept proposed in ACL 2025, offering a viable solution for construction recognition in low-resource scenarios—particularly applicable to the widespread challenges of "scarce annotated data and limited computational resources" in linguistic research.

To further investigate the systematic biases of large language models in identifying ditransitive constructions, this study conducted a sample analysis of the errors made by Qwen3-MAX on the test set. It was found that its misjudgments primarily clustered in the following types of peripheral or complex syntactic environments. These cases reveal that the model's judgment of form-meaning pairings remains excessively influenced by surface-level syntactic sequences.

The first type is misjudgment of ditransitive constructions in subordinate clauses. The model failed to accurately identify ditransitive structures nested within subordinate clauses. For example, in the sentence "If Odin does not bring them peace and good harvests, then it is the duty of God and the King.", the model correctly recognized that "bring them peace and good harvests" conforms to the "S-V-IO-DO" form but overlooked that this structure occurs within a conditional adverbial clause and is not an independent main clause. This reflects the model's insufficient sensitivity to the contextual syntactic status of constructions and an over-reliance on local sequential patterns.

Another category is confusion between idiomatic and literal ditransitive meanings. The model failed to distinguish ditransitive expressions with idiomatic meanings from those with literal transfer meanings. For instance, in "I'll give her a call and find out.", the model judged "give her a call", meaning to make a phone call, as a typical ditransitive construction, ignoring that this expression has solidified into an idiom whose semantics is not typical transfer of possession but rather transfer of action. Such errors indicate that the model's understanding of semantic roles is still influenced by the surface frequency of verb-noun collocations and fails to fully integrate the conventionalized meaning of the construction within specific contexts.

The last one is over-identification of ditransitives in non-finite structures. The model erroneously incorporated verb-double object structures within non-finite phrases into the main clause construction judgment. For example, in "Seven ECOWAS heads of state [...] called for a ceasefire [...], giving ECOMOG carte blanche to impose it by force.", the model identified the present participle phrase "giving ECOMOG carte blanche" as an independent ditransitive construction, even though this structure is syntactically dependent on the main clause and does not constitute an independent predicate core. This suggests a fuzziness in the model's judgment of syntactic hierarchy and a failure to strictly distinguish between predicate structures in the main clause and those in modifying elements.

These error cases collectively reveal a key limitation of the large model in ditransitive construction identification: the recognition of constructions remains highly dependent on matching the surface S-V-N-N sequence, while its judgment capabilities regarding syntactic hierarchy, semantic conventionality, and contextual dependence are weak. In contrast, the LoRA-fine-tuned model, through exposure to more diverse contextual annotations during training, demonstrated a stronger ability to distinguish such marginal cases. This further confirms that targeted data training can enhance the model's sensitivity to the semantic and pragmatic dimensions of form-meaning pairings in constructions.

Although LoRA fine-tuning significantly improved the model's ability to judge the core semantics and complex forms of the ditransitive construction, a small number of errors remained on the test set with a accuracy 0.936. Analysis of these residual errors reveals that after targeted training, the model's limitations are mainly concentrated in the fine-grained discrimination of non-canonical syntactic boundaries and argument relations highly dependent on context.

The first sort is overgeneralization of missing required argument. The model occasionally misclassified nontransitive structures carrying only one object as ditransitive. For example, in the sentence "I earned a few sous so I became more fantastical...", the model annotated "earned a few sous" as a ditransitive construction. However, this structure only has a theme, a few sous, and completely lacks the recipient, a required argument, thus not conforming to the core argument framework of "agent-recipient-theme" for the ditransitive construction. This indicates that in individual cases, the model's suppression of overgeneralization based on a verb's strong ditransitive tendency, such as earn can appear in ditransitive contexts, was insufficient.

The second type is misunderstanding of elements within complex prepositional structures. When analyzing sentences embedded with complex prepositional phrases, the model may incorrectly bind the prepositional object as the recipient of a ditransitive. For instance, in "I had some petty cash with which to buy them all sandwiches...", the model judged "had some petty cash" as ditransitive and considered them to be the recipient. However, syntactic analysis shows that "them" is actually part of the subsequent prepositional phrase "to buy them all sandwiches..." and has no direct argument relation with the main clause verb "had." Such errors reflect the challenges the model faces in performing precise syntactic role assignment in sentences with long-distance syntactic dependencies and fragmented argument structures.

The last one is misjudgment of the syntactic status of constructions in nested clauses. Similar to the baseline model, the fine-tuned model, in very rare cases, still misjudges constructions within nested clauses, but the nature of the error shifts from "failure to recognize" to "status misjudgment." For example, in "...provided one gives him no work to do.", the model correctly identified "gives him no work" as a ditransitive structure but may have overlooked that this structure is within a conditional adverbial clause, requiring special consideration regarding its syntactic status under the overall annotation guidelines. This suggests that while the model's sensitivity to the syntactic hierarchy of constructions has been enhanced, there is still room for improvement in the most nuanced contextual discrimination.

Comparing the error patterns of the fine-tuned model and the baseline model Qwen3-MAX clearly reveals the path of improvement and the remaining challenges. That is how to shift from form-driven to semantics-driven. The errors of the baseline model mostly stemmed from the mechanical matching of surface S-V-N-N sequences, like misjudging the idiom *give a call*, which is a typical manifestation of the dissociation of form and meaning. In contrast, the residual errors of the fine-tuned model occur more in the fine-grained discrimination of argument relations in complex syntactic environments, indicating that it has preliminarily established a judgment framework centered on "CAUSE-RECEIVE" semantics with formal constraints.

The nature of the remaining challenge lies in the fact that the errors of the fine-tuned model are primarily related to syntactic complexity and contextual depth, rather than a misunderstanding of the construction's core semantics. This points to a key direction for future research. To further push the ceiling of construction identification, it is necessary not only to have more data but also to incorporate the modeling of explicit linguistic knowledge, such as syntactic tree structures and argument dependency relations, into the training process.

This analysis indicates that LoRA fine-tuning effectively guides the model from "guessing based on surface correlations" to "judging based on the semantic core of the construction." However, to achieve human-expert-like robustness, continued exploration into the deep integration of structured linguistic knowledge is still required.

### 5.2 Theoretical and Practical Contributions

This study's theoretical innovation is multi-faceted and particularly noteworthy in several key respects. Firstly, it represents a significant departure from the conventional research paradigm that has long been dominated by the widespread use of BERT. In the existing body of research, BERT has been the go-to model for numerous natural language processing tasks, and its influence has permeated various subfields, including those related to grammatical construction identification. However, this study boldly breaks away from this norm. Instead of relying on the well-established but perhaps somewhat over-relied-upon BERT, it explores new avenues and alternative approaches, thereby injecting fresh perspectives and methodologies into the field. This shift not only challenges the status quo but also opens up new possibilities for more effective and tailored solutions in construction identification research.

Secondly, an even more remarkable aspect of this study's theoretical innovation is its employment of a LLM originating from China. In the global landscape of large language models, the majority of the prominent and widely studied ones have predominantly come from Western countries or international tech giants. This has led to a certain imbalance in research focus and a lack of exploration into models with diverse cultural and linguistic backgrounds. In the specific context of construction identification, there has been a conspicuous absence of studies centered around Chinese large language models. This study fills this critical gap in the literature. By utilizing a Chinese LLM, it not only broadens the scope of research in this area but also takes into account the unique linguistic characteristics, cultural nuances, and grammatical structures inherent in the Chinese language. This, in turn, enables a more comprehensive and in-depth understanding of construction identification across different languages and cultural contexts, contributing to the development of a more universal and inclusive theoretical framework for this field.

By translating Goldberg's (1995) form-meaning pairing theories into annotatable, learnable task objectives, the model achieved precise acquisition of double-object construction prototypical features through 5,000 samples, addressing the core debate on "whether construction theories can be scaled for implementation" (Doumen et al., 2024).

In summary, this study's theoretical innovation lies not only in its bold break from the BERT-centric research norm but also in its pioneering use of a Chinese LLM, which addresses a long-standing gap in construction identification research and paves the way for future cross-cultural and multilingual studies in this domain.

Practically, the highly significant difference ($p = 8.38e-12$) between the RAG system (pure theory) and the fine-tuned model, theory plus instances, demonstrates that construction recognition requires "theoretical constraints to ensure semantic correctness and instance learning to ensure formal generalization." A single theoretical knowledge base cannot cover form variants in language use, providing clear methodological guidance for future computational construction grammar research.

The fine-tuned Qwen3-8B achieved an accuracy of 0.936 with an error rate of only 6.4%, approaching human annotator consistency levels 84%-90% (Bybee, 2010), at a total cost of just 18 RMB. Compared to international models like GPT-3, its localization advantages are significant, offering a viable alternative for resource-constrained research teams.

Instances labeled as 1 account for 19% of the dataset, while those labeled as 0 account for 81%, potentially improve the model's sensitivity to double-object constructions. This fits with the real-world need in construction recognition for there a relatively low ratio of double-object instances in the whole sentences. The sensitivity to double-object construction can enhance the capability dealing with raw corpora.

### 5.3 Limitations and Future Directions

Reliance solely on the BN corpus lacks coverage of American English, social media variants, and fails to account for distributional differences in double-object constructions across genres, such as lower frequency in argumentative texts than in spoken language, leaving the model's cross-genre generalization performance unverified.

Despite its prevalence, the fine-tuning paradigm faces critical limitations. It is inherently data-dependent, and creating high-quality, annotated datasets for linguistic constructions requires significant expert labor, a challenge for low-resource phenomena (Plank, 2016). Moreover, the process is largely a black box; the linguistic knowledge acquired is diffusely encoded across millions of parameters, making the model's reasoning opaque and difficult to verify against linguistic principles (Rudin, 2019). Crucially, a fine-tuned model is statically knowledge-bound. Its knowledge is frozen at training, incapable of integrating new linguistic insights without costly retraining, which is a major drawback for dynamic research fields.

The RAG system's failure to integrate knowledge graph technology prevents structured representation of "agent-verb-recipient-object" relationship networks. Additionally, the lack of an instance base hinders handling of low-frequency verb constructions and marginal semantic cases. The RAG in this paper lack the ability to retrieve and synthesize information from structured linguistic resources, like FrameNet (Ruppenhofer et al. 2006). This research defines semantic roles for ditransitive constructions, like Recipient, Theme. What is more the large-scale corpor, such as BNC/COCA, can provide authentic ditransitive instances for analogy, which makes it a powerful tool for grounding constructional analysis in explicit knowledge.

There are some suggestions for future research. The first direction is data and knowledge base optimization. Supplement with multi-source corpora such as COCA (Corpus of Contemporary American English) and social media, balance class proportions (1:1), and construct a three-tier knowledge base combining theoretical literature plus typical instances plus knowledge graphs. Adopt a text retrieval plus graph retrieval dual-layer mechanism to enhance recognition performance for low-frequency constructions and complex forms.

Another choice should be low-resource and cross-linguistic expansion. Drawing on Few-shot learning concept, explore model fine-tuning effects with minimal annotated data like 500 samples to reduce annotation costs. Extend the binary classification framework to Chinese double-object constructions to verify the model's cross-linguistic transferability, providing computational empirical support for cross-linguistic construction grammar theories.

Last but not least is the deep integration with teaching applications. Based on Chinese students' acquisition difficulties, optimize the model's output format to not only provide labeling results but also error type annotations, such as preposition redundancy, argument position reversal, and correction suggestions, developing a dedicated tool for second language teaching.

## 6 Conclusion

This study systematically verified the recognition performance of LoRA fine-tuning, native large models, and RAG systems using 5,000 English double-object construction binary classification samples from the BNC corpus, employing McNemar

tests + Bonferroni correction for statistical validation. The core conclusions are as follows:

The fine-tuned Qwen3-8B achieved an accuracy of 0.936, significantly outperforming Qwen3-MAX (0.898, p=0.0134) and RAG (0.792, p=8.38e-12). Its advantage stems from precise acquisition of double-object constructions form-meaning dual criteria.

Qwen3-MAX's accuracy significantly exceeded RAG's (p=3.88e-07), proving that native large models possess foundational construction recognition capabilities but struggle to overcome semantic constraint and complex form recognition bottlenecks without targeted fine-tuning.

The performance in this research is fine-tuned performs better than Qwen3-MAX, and that of RAG is worst. This result is statistically significant and highly consistent with construction grammar theories and RAG technology evolution trends, validating the scientific of the fine-tuning pathway.

This study constructed a lightweight construction recognition solution, filling the application gap of domestic large models in specialized double-object construction binary classification tasks. It also provides a complete example of how to shift from theory implementation to statistical validation and then to practical application for computational construction grammar research. Future advancements through data optimization, knowledge base upgrades, and ross-linguistic expansion will further promote the deep integration of construction grammar theories and large models, offering more valuable technical support for linguistic research and second language teaching